\documentclass[final]{cvpr}

\usepackage{times}
\usepackage{epsfig}
\usepackage{graphicx}
\usepackage{amsmath}
\usepackage{amssymb}
\usepackage{enumitem}
\usepackage{color}
\usepackage{array,hhline}
\usepackage{booktabs} 
\usepackage{subfigure}

\usepackage{multirow}

\usepackage[pagebackref=true,breaklinks=true,colorlinks,bookmarks=false]{hyperref}

\def\cvprPaperID{4572} 
\def\confYear{CVPR 2021}

\begin{document}

\title{Instant-Teaching: An End-to-End Semi-Supervised \\ 
Object Detection Framework}

\newcommand*{\affmark}[1][*]{\textsuperscript{#1}}
\author{%
Qiang Zhou\affmark[],~~~~ Chaohui Yu\affmark[],~~~~ Zhibin Wang\affmark[],~~~~ Qi Qian\affmark[],~~~~ Hao Li\affmark[]\\
Alibaba Group\\
\{jianchong.zq,~huakun.ych,~zhibin.waz,~qi.qian,~lihao.lh\}@alibaba-inc.com\\
}

\maketitle

\pagestyle{empty}
\thispagestyle{empty}

\begin{abstract}

Supervised learning based object detection frameworks demand plenty of laborious manual annotations, which may not be practical in real applications. Semi-supervised object detection (SSOD) can effectively leverage unlabeled data to improve the model performance, which is of great significance for the application of object detection models. In this paper, we revisit SSOD and propose Instant-Teaching, a 
completely end-to-end and effective SSOD framework, which uses instant pseudo labeling with extended weak-strong data augmentations for teaching during each training iteration. To alleviate the confirmation bias problem and improve the quality of pseudo annotations, we further propose a co-rectify scheme based on Instant-Teaching, denoted as Instant-Teaching$^*$.
Extensive experiments on both MS-COCO and PASCAL VOC datasets substantiate the superiority of our framework.
Specifically, our method surpasses state-of-the-art methods by 4.2 mAP on MS-COCO when using $2\%$ labeled data. Even with full supervised information of MS-COCO, the proposed method still outperforms state-of-the-art methods by about 1.0 mAP. On PASCAL VOC, we can achieve more than 5 mAP improvement by applying VOC07 as labeled data and VOC12 as unlabeled data.

\end{abstract}


\section{Introduction}

Deep neural networks~\cite{krizhevsky2012imagenet,simonyan2014very,he2016deep} have significantly improved the performance of diverse computer vision applications, \eg, image classification and object detection. In order to avoid overfitting and achieve better performance, a large amount of accurate human-annotated data is needed to train a deep learning model. However, the assumption of having a sufficient amount of accurate labeled data for training may not hold, especially for object detection tasks, which need annotations with accurate class labels and precise bounding box coordinates. Thus, a natural idea is to leverage abundant unlabeled data to facilitate learning in the original task. To relax the dependency of manually labeled data, a promising approach is called semi-supervised learning (SSL)~\cite{chapelle2010semi}.

SSL has recently received increasing attention from the community, since it provides effective methods of using unlabeled data to facilitate model learning with limited annotated data. Most of the existing SSL methods focus on image classification tasks and there are multiple strategies for semi-supervised learning, \eg, self-training~\cite{scudder1965probability,xie2020self} and co-training~\cite{blum1998combining,qiao2018deep}. Recently, one popular line of research uses consistency losses for semi-supervised learning~\cite{lee2013pseudo,rasmus2015semi,laine2016temporal,tarvainen2017mean,miyato2018virtual,sajjadi2016regularization,tarvainenweight,xie2019unsupervised,berthelot2019mixmatch,berthelot2019remixmatch,sohn2020fixmatch}. They either adopt ensemble learning algorithms to enforce the predictions of the unlabeled data to be consistent across multiple models, or constrain the model predictions to be invariant to noise. Another popular line of SSL research focuses on more effective data augmentations to improve the generalization and robustness of the model, in which some learning-based and more complex data augmentation strategies \cite{berthelot2019remixmatch,xie2019unsupervised,cubuk2019randaugment,berthelot2019mixmatch,sohn2020fixmatch} greatly improve the performance of SSL on image classification tasks.

\begin{figure*}[t!]
	\begin{center}
	\includegraphics[width=0.95\textwidth]{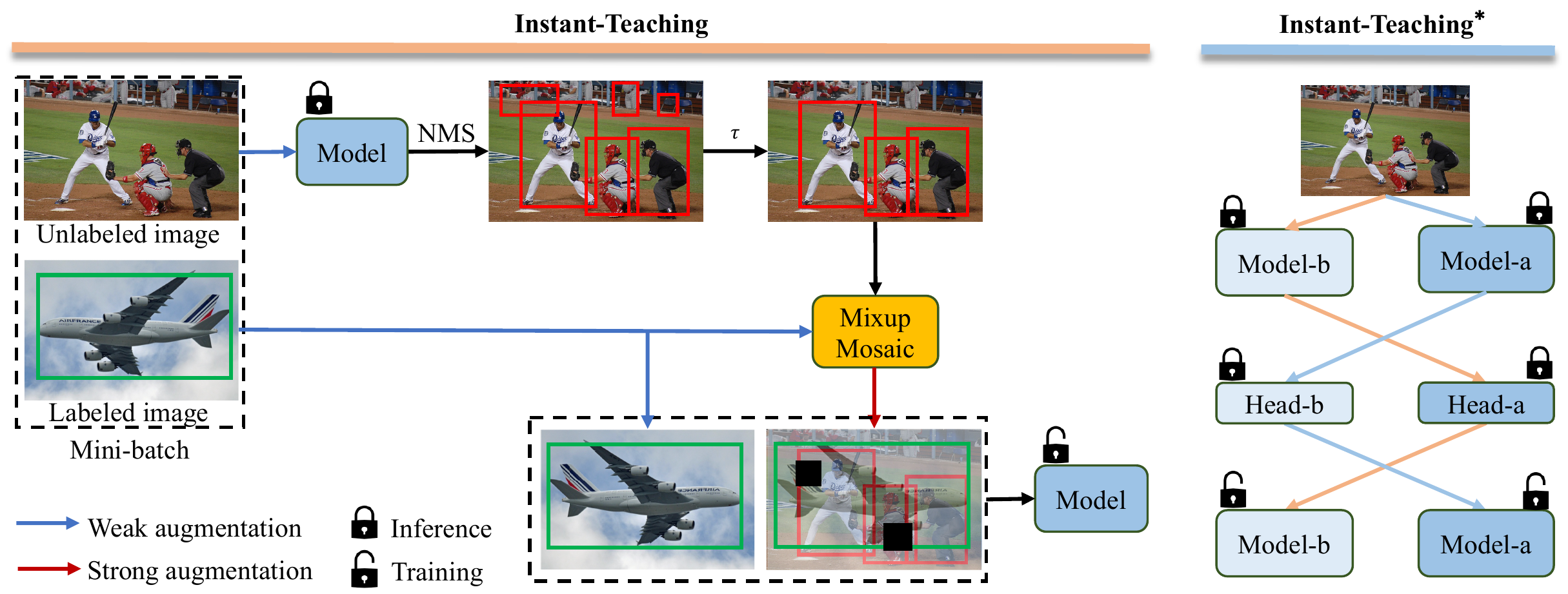}
	\end{center}
	\vspace{-.1in}
	\caption{The proposed semi-supervised object detection framework. Instant-Teaching includes instant pseudo labeling with extended weak-strong data augmentations. Instant-Teaching$^*$ represents Instant-Teaching combined with our co-rectify scheme.}
	\label{fig:ssl_framework}
	\vspace{-.1in}
\end{figure*}

Although semi-supervised learning has made great progress in the field of image classification, there is a paucity of literature focus on semi-supervised object detection (SSOD). 
The recently proposed STAC~\cite{sohn2020simple} performs best among existing SSOD methods and outperforms the supervised model by a large margin, which is of great significance to the research of SSOD.
However, we find that STAC still has some problems. First, its training procedure is complicated and inefficient. Before model training, STAC needs to train a teacher model, and then it uses the teacher model to pre-generate pseudo annotations of unlabeled data.
Second, during model training, the pre-generated pseudo annotations will no longer be updated, and the constant label will limit its performance.
In this paper, to address the above two problems, we propose a novel end-to-end SSOD framework, Instant-Teaching, which uses instant pseudo labeling with extended weak-strong data augmentations for teaching during each training iteration. Specifically, as shown in Fig.~\ref{fig:ssl_framework}, during each training iteration, Instant-Teaching will first generate pseudo annotations of unlabeled data with weak data augmentations in a mini-batch, and then the predicted annotations will instantly be used as the ground-truth of the same image with strong data augmentations for training. 
The advantage of Instant-Teaching is that as the model converges during training, the quality of pseudo annotations will be improved instantly. 
The weak-strong data augmentation scheme is inherited from STAC, which has been proven to be effective in combination with pseudo annotations, and we further extend the strong data augmentations to include Mixup and Mosaic.
In addition, the confirmation bias~\cite{tarvainen2017mean} is a common problem in SSL. To alleviate this issue, we further propose a co-rectify scheme based on Instant-Teaching, denoted as Instant-Teaching$^*$. Instant-Teaching$^*$ simultaneously trains two models that have the same structure but share different weights and these two models help each other to rectify false predictions. During inference, we still only use a single model so that it does not increase inference time.

We test the efficacy of Instant-Teaching$^*$ on PASCAL VOC~\cite{everingham2010pascal} and MS-COCO~\cite{lin2014microsoft} datasets, and follow the experimental protocols used in the latest state-of-the-art SSOD literature STAC~\cite{sohn2020simple} to evaluate the performance.
It is worth mentioning that, our Instant-Teaching$^*$ framework outperforms state-of-the-art methods at all experimental protocols, and achieves state-of-the-art performance on semi-supervised object detection learning.

The contributions of this paper are as follows:
\begin{itemize}

\item We propose a novel SSOD framework, called Instant-Teaching, which uses instant pseudo labeling with extended weak-strong data augmentations for teaching during each training iteration. Instant-Teaching is an end-to-end framework and can effectively leverage the unlabeled data.

\item To alleviate the confirmation bias problem and improve the quality of pseudo annotations, we further propose a co-rectify scheme based on Instant-Teaching, denoted as Instant-Teaching$^*$.

\item Our extensive experiments on PASCAL VOC and MS-COCO datasets demonstrate the significant efficacy of our Instant-Teaching$^*$ framework. 
\end{itemize}

\section{Related Work}
\noindent \textbf{Object detection} is an important computer vision task and has received considerable attention in recent years~\cite{girshick2014rich,girshick2015fast,ren2015faster,he2017mask,lin2017feature,cai2018cascade,hu2018relation,liu2016ssd,redmon2016you,lin2017focal}. One line of research focuses on strong two-stage object detectors~\cite{girshick2014rich,girshick2015fast,ren2015faster,dai2016r,he2017mask,lin2017feature,cai2018cascade,hu2018relation}, which first generate a sparse set of regions of interest (RoIs) with a Region Proposal Network (RPN), and then perform classification and bounding box regression. Another line of research develops fast single-stage object detectors~\cite{liu2016ssd,redmon2016you,lin2017focal,law2018cornernet,duan2019centernet,tian2019fcos}. However,
these methods train stronger or faster models on a large amount of accurate human-annotated data, which is expensive and time-consuming to acquire in real applications. In this work, we follow the popular two-stage object detector (Faster-RCNN~\cite{ren2015faster}) to develop our framework. Different from previous methods training models only on labeled data, we train our object detector on both labeled and unlabeled data with our proposed semi-supervised learning strategy.

\noindent \textbf{Semi-supervised learning (SSL)} exploits the potential of unlabeled data to facilitate model learning with limited annotated data. Most of the existing SSL methods focus on image classification tasks and most of the works~\cite{berthelot2019mixmatch,xie2019unsupervised,miyato2018virtual,sajjadi2016regularization,laine2016temporal,tarvainen2017mean,berthelot2019remixmatch,sohn2020fixmatch} are consistency-based methods, which constrain the model to be invariant to the noise.
Pseudo labeling based methods \cite{lee2013pseudo,bachman2014learning,iscen2019label,arazo2020pseudo,xie2020self} improve the performance of SSL by generating high-quality hard labels (\ie, the $\mathop{\arg\max}$ of the output class probability) of unlabeled data with a predefined threshold and retraining the model. 
Recently, data augmentations have proven to be a powerful paradigm for boosting SSL on image classification~\cite{berthelot2019mixmatch,xie2019unsupervised,cubuk2019randaugment,berthelot2019remixmatch,sohn2020fixmatch}.
MixMatch~\cite{berthelot2019mixmatch} improves SSL by guessing low-entropy labels for data-augmented unlabeled data and mixes labeled and unlabeled data using Mixup. FixMatch~\cite{sohn2020fixmatch}, UDA~\cite{xie2019unsupervised} and ReMixMatch~\cite{berthelot2019remixmatch} have shown that RandAugment~\cite{cubuk2019randaugment} and CTAugment~\cite{berthelot2019remixmatch} can significantly facilitate learning of SSL.

\noindent \textbf{Semi-supervised object detection (SSOD)} applies semi-supervised learning to object detection. Recently, a few existing works~\cite{misra2015watch,tang2016large,gao2019note,jeong2019consistency,tang2020proposal,li2020improving,sohn2020simple,jeong2020interpolation} propose to train object detectors on both labeled data and unlabeled data by incorporating SSL into object detection. The methods in~\cite{misra2015watch,tang2016large} depend on additional context (\eg, temporal information from video).
The method in~\cite{radosavovic2018data} proposes data distillation to automatically generate new training annotations by ensembling predictions of multiple transformations of unlabeled data. NOTE-RCNN~\cite{gao2019note} proposes to iteratively perform bounding box mining and detector retraining. S$^4$OD~\cite{li2020improving} proposes a selective net as a heuristic for selecting bounding boxes to improve object detection with unlabeled web images.
CSD~\cite{jeong2019consistency} proposes a consistency-based SSL method for object detection, which uses flip augmentation and consistency constraints to enhance detection performance.
Based on CSD, ISD~\cite{jeong2020interpolation} proposes to use interpolation regularization to further improve the performance of SSL for object detection.
Recently, STAC~\cite{sohn2020simple} develops a SSL framework for object detection that combines self-training and consistency regularization based on strong data augmentations, which achieves state-of-the-art results. 


Inspired by these methods, this paper exploits the effective usage of pseudo annotations as well as data augmentations and co-rectify scheme to further improve the performance of SSL for object detection in a more efficient and simpler way.

\section{Method}

In this section, we first give the problem definition of our semi-supervised object detection task (see Section~\ref{method:probdef}). Then, we show an overview of our Instant-Teaching$^*$ framework (see Section~\ref{method:tof}), which consists of instant pseudo labeling with extended weak-strong data augmentations and co-rectify scheme (see Section~\ref{method:instant-teaching}).

\subsection{Problem definition}
\label{method:probdef}
In semi-supervised object detection (SSOD), we are given a set of labeled data $\mathcal{D}_{l}=\{(\mathbf{x}^{l}_{i},y^{l}_{i})\}^{n_l}_{i=1}$ and a set of unlabeled data $\mathcal{D}_{u}=\{\mathbf{x}^{u}_{j}\}^{n_u}_{j=1}$, where $\mathbf{x}$ and $y$ denote image and ground-truth annotations (class labels and bounding box coordinates) respectively. The goal of SSOD is to train object detectors on both labeled and unlabeled data.

\subsection{The overview framework}
\label{method:tof}

As shown in Fig.~\ref{fig:ssl_framework}, our Instant-Teaching$^*$ framework is mainly composed of two modules, namely, instant pseudo labeling with weak-strong data augmentations and co-rectify.
It is worth mentioning that, the first module of instant pseudo labeling with weak-strong data augmentations already forms a complete SSOD framework, denoted as Instant-Teaching, which also outperforms state-of-the-art methods.

These two modules have their own focus, among which instant pseudo labeling with weak-strong data augmentations enables our method to be trained end-to-end, and the quality of pseudo annotations is instantly improved as the model converges.
Moreover, weak-strong data augmentations enforce the model to maintain consistent predictions between the weakly augmented and the strongly augmented unlabeled data. In this way, the model can learn useful information from the pseudo annotations generated by itself. The co-rectify scheme trains two models with the same structure simultaneously and these two models help each other to rectify false predictions, thus alleviating the common confirmation bias problem and further improving the model performance.

Please note that although our Instant-Teaching$^*$ trains two models at the same time, we only use a single model (Model-a) during inference, which does not increase inference time.

\subsection{Instant-Teaching$^*$}
\label{method:instant-teaching}

\paragraph{Instant pseudo labeling.} It is beneficial to update the pseudo annotations with a more precise model during the training process, which motivates us to propose instant pseudo labeling. 
Instant pseudo labeling performs model training and pseudo-label generation at the same time, which is end-to-end and different from the latest STAC~\cite{sohn2020simple} framework. STAC needs to train a teacher model in advance to generate pseudo annotations of unlabeled data. Moreover, STAC does not update the generated pseudo annotations during training, which limits its performance.

To be more specific, we decompose each training iteration into two steps. In the first step, we use the current model to generate pseudo annotations of unlabeled data in a mini-batch. Note that in this step, we apply weak augmentations $\alpha(\cdot)$ to unlabeled data (unless otherwise specified, we only use random flip as weak augmentation in all experiments). In the second step, we apply strong augmentations $A(\cdot)$ to the same unlabeled data with pseudo annotations generated in the first step, and update the model parameters with a entire training objective, which consists of a supervised loss and an unsupervised loss. Note that in this step, to get a fair comparison with STAC, we only apply strong data augmentations to unlabeled data, while still applying weak augmentations to labeled data.
In fact, the performance of the model will be relatively poor in the initial training phase. In order to guarantee the quality of the generated pseudo annotations, we always apply non-maximum suppression (NMS) and confidence-based box filtering with a high threshold $\tau$ in the first step (unless otherwise specified, we use $\tau=0.9$ in all experiments).

Overall, the model is trained by jointly minimizing the supervised loss and unsupervised loss as follows:
\begin{equation}
	\label{eq:loss_total}
	\ell = \ell_s + \lambda_u \ell_u,
\end{equation}
where we use $\lambda_u$ to balance the supervised loss $\ell_s$ and the unsupervised loss $\ell_u$.

The supervised loss $\ell_s$ consists of a classification loss $L_{cls}$ (a standard cross-entropy loss), and a bounding box regression loss $L_{reg}$ (a $L_1$ loss). $\ell_s$ can be computed as:
\begin{equation}
	\label{eq:loss_sup}
	\begin{aligned} \ell_s =& \sum_{l} [ \frac{1}{N_{cls}}\sum_{i}L_{cls}(p(c_i \mid \alpha(\mathbf{x}_l)),c_{i}^*) \\ &+\frac{\lambda}{N_{reg}}\sum_{i}c_{i}^*L_{reg}(p(\mathbf{t}_i \mid \alpha(\mathbf{x}_l)), \mathbf{t}_i^*) ].
	\end{aligned}
\end{equation}
In the above equation, $l$ is the index of labeled images in a mini-batch, $i$ is the index of an anchor in a single image, $p(c_i \mid \mathbf{x})$ is the predicted probability of anchor $i$ being an object in image $\mathbf{x}$, $p(\mathbf{t}_i \mid \mathbf{x})$ is the 4-dimensional coordinates of an predicted bounding box, $c_{i}^*$ and $\mathbf{t}_i^*$ are the human-annotated ground-truth class label and bounding box coordinates respectively.

When computing the unsupervised loss $\ell_u$, we first compute the model's predicted class probability distribution and box coordinates based on weakly augmented unlabeled data in a mini-batch: $(c^u, \mathbf{t}^u) = p(c, \mathbf{t} \mid \alpha(\mathbf{x}_u))$. Then we use the hard label $\hat{c}^u = \mathop{\arg\max}(c^u)$ as the final class label of pseudo annotations. In addition, the unsupervised loss is computed on strongly augmented unlabeled data and can be written as:
\begin{equation}
	\label{eq:loss_uns}
	\begin{aligned} \ell_u =& \sum_{u} [ \frac{1}{N_{cls}}\sum_{i}L_{cls}(p(c_i \mid A(\mathbf{x}_u)),\hat{c}_{i}^u) \\
	&+\frac{\lambda}{N_{reg}}\sum_{i} (\text{max}(c_i^u) \geq \tau ) L_{reg}(p(\mathbf{t}_i \mid A(\mathbf{x}_u)), \mathbf{t}_i^u) ],
	\end{aligned}
\end{equation}
where $u$ is the index of unlabeled images in a mini-batch, $\hat{c}_{i}^u$ and $\mathbf{t}_i^u$ are pseudo annotations of unlabeled data generated by the model itself, and $\tau$ is the confidence threshold.

\begin{figure}[t!]
	\begin{center}
	\includegraphics[width=0.33\textwidth]{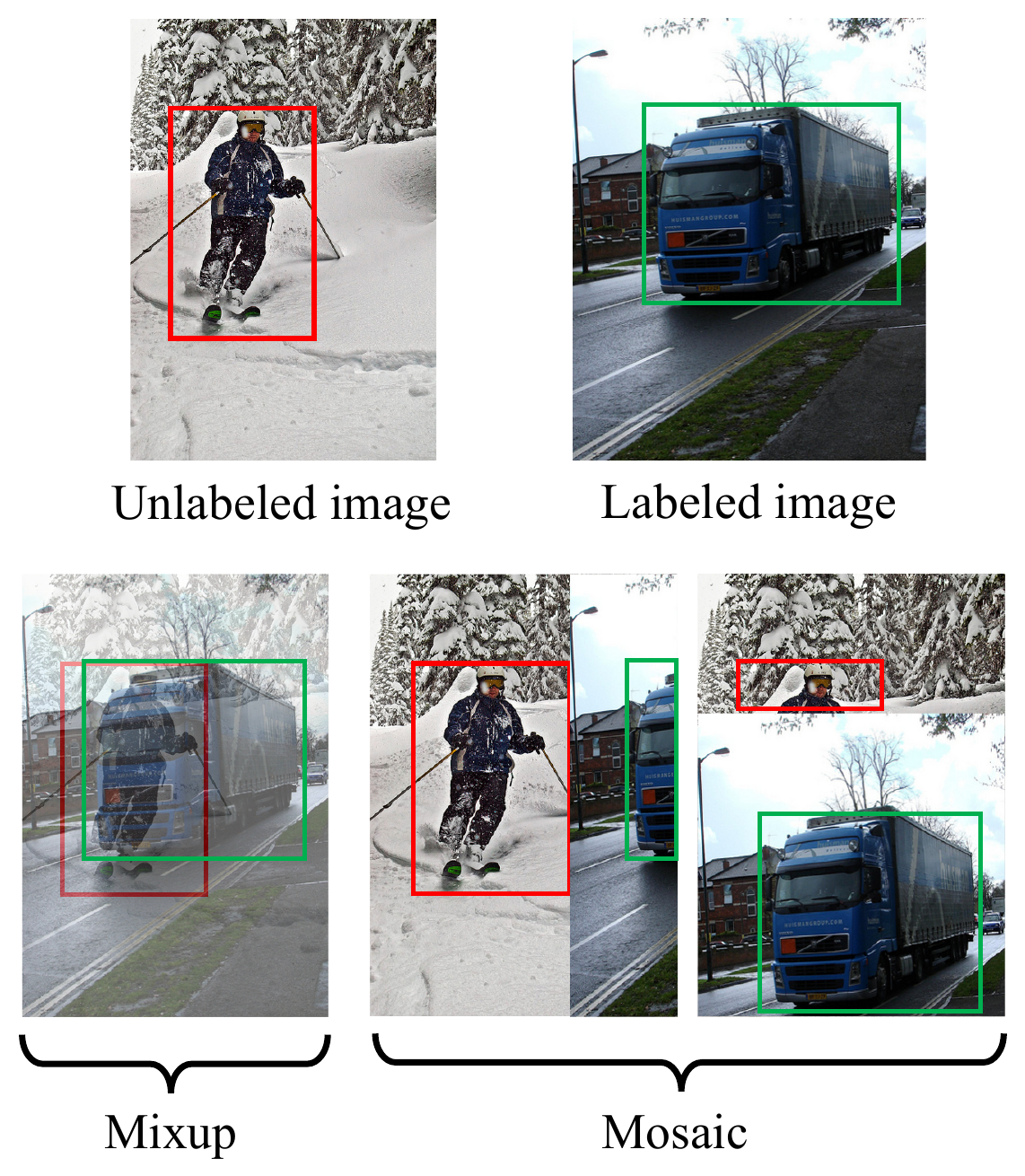}
	\end{center}
	\vspace{-.15in}
	\caption{Mixup and Mosaic data augmentations for semi-supervised object detection learning.}
	\label{fig:instance_mixup}
	\vspace{-.2in}
\end{figure}

\paragraph{Weak-strong data augmentations.} How to encourage the model to learn useful information from the pseudo annotations generated by the model itself is essential to all self-training based SSL methods. Weak-strong data augmentation scheme is a promising practice, which has been proven in semi-supervised image classification~\cite{sohn2020fixmatch} and semi-supervised object detection~\cite{sohn2020simple}. Weak-strong data augmentations enforce the model to maintain consistent predictions between the weakly augmented and the strongly augmented unlabeled data, and thus encourage the model to learn useful information from the pseudo annotations.

Intuitively, the key of weak-strong data augmentation scheme lies in the difference between weak augmentations and strong augmentations. When the weak augmentations remain unchanged, the more complex and appropriate the strong augmentations, the more information the model can learn from the pseudo annotations.
Based on this hypothesis, we extend the strong data augmentations of STAC and introduce more complex augmentations for unlabeled data, including Mixup~\cite{zhang2017mixup} and Mosaic~\cite{bochkovskiy2020yolov4}. The experimental results also reveal that our extended weak-strong data augmentations can further improve the performance of semi-supervised object detection.

Specifically, as shown in Fig.~\ref{fig:instance_mixup} (Mixup), given an unlabeled image $\mathbf{x}_u$ and its pseudo annotations ($b_u$, $c_u$), where $b_u$ are the 4-dimensional box coordinates and $c_u$ are the one-hot class labels of these pseudo boxes (note that we use hard label when the confidence score is larger than the confidence threshold $\tau$). We first randomly choose one labeled image $\mathbf{x}_l$ with ground-truth annotations ($b_l$, $c_l$) from the mini-batch. Next, we mix these two images and their one-hot labels and bounding box coordinates with a mixing coefficient $\lambda_m$ drawn from the $Beta(\alpha_{m}, \alpha_{m})$ distribution, where $\alpha_{m}=1.0$. Finally, we use the mixed image and soft class labels and bounding box coordinates to substitute the image content and pseudo annotations of the unlabeled image $\mathbf{x}_u$, which can be computed as:
\begin{equation}
\label{eq:mixup}
\left\{
     \begin{array}{lcl}
     \lambda_m &\sim& Beta(\alpha_{m}, \alpha_{m}), \\
     \mathbf{x}_u &=& \lambda_m \mathbf{x}_u + (1 - \lambda_m) \mathbf{x}_l, \\
     c_u &=& \lambda_m c_u \cup (1 - \lambda_m) c_l, \\ 
     b_u &=& b_u \cup b_l. 
     \end{array}
\right.
\end{equation}

As for Mosaic, as shown in Fig.~\ref{fig:instance_mixup} (Mosaic), given an unlabeled image $\mathbf{x}_u$ and a labeled image $\mathbf{x}_l$ in a mini-batch, we randomly perform two kinds of mixing styles (horizontal mixing and vertical mixing) and mix their corresponding annotations accordingly.
By applying Mixup and Mosaic data augmentations to unlabeled data, we can improve the model robustness to pseudo annotation noise and alleviate the overfitting problem in model training.

Note that, for a fair comparison, we only perform Mixup and Mosaic data augmentations on unlabeled data and keep labeled data with weak data augmentation unchanged in all our experiments, which is the same as STAC~\cite{sohn2020simple}.

\paragraph{Co-rectify.} 

Confirmation bias~\cite{tarvainen2017mean} is a common problem in semi-supervised learning. When the model generates incorrect predictions with high confidence, these incorrect predictions will be further strengthened 
through incorrect pseudo annotations.
In other words, the model itself is difficult to rectify these false predictions.

To alleviate this problem, we propose a co-rectify scheme, which trains two models $f_a(\cdot)$ (Model-a) and $f_b(\cdot)$ (Model-b) simultaneously. These two models help each other to rectify the false predictions, as shown in Fig.~\ref{fig:ssl_framework}. The key to the success of co-rectify is that the two models will not converge to the same model. We take two measures to ensure that the two models converge independently. First, although the two models have the same structure, they use different initialization parameters. Second, although the two models share the same data in each mini-batch, their data augmentations and pseudo annotations are also different.

We take model $f_a(\cdot)$ as an example and the rectified pseudo annotations of model $f_b(\cdot)$ are constructed in a similar way. When generating pseudo annotations during each training iteration, model $f_a(\cdot)$ first predicts class probabilities $c_i$ and bounding box coordinates $\mathbf{t}_i$ on the weakly augmented unlabeled image $\mathbf{x}_u$.
Then, we use the detection head in model $f_b(\cdot)$ to refine the class probabilities $c_i^r$ and bounding box coordinates $\mathbf{t}_i^r$ by taking the predicted boxes $\mathbf{t}_i$ as proposals.
Finally, the rectified class probabilities $c_i^*$ are averaged from $c_i$ and $c_i^r$,  and the rectified bounding box coordinates $\mathbf{t}_i^*$ are the weighted average of $\mathbf{t}_i$ and $\mathbf{t}_i^r$. The co-rectify process can be computed as:
\begin{equation}
\label{eq:co-rectify}
\left\{
	\begin{array}{ccl}
	(c_i, \mathbf{t}_i) &=& f_a(\mathbf{x}_u), \\
	(c_i^r, \mathbf{t}_i^r) &=& f_b(\mathbf{x}_u; \mathbf{t}_i), \\
	c_i^* &=& \frac{1}{2}(c_i + c_i^r), \\
	\mathbf{t}_i^* &=& \frac{1}{c_i + c_i^r}(\mathbf{t}_i c_i + \mathbf{t}_i^r c_i^r).
	\end{array}
\right.
\end{equation}

\section{Experiments}

We test our proposed semi-supervised object detection framework Instant-Teaching$^*$ on the large-scale dataset MS-COCO~\cite{lin2014microsoft} and report the mAP over 80 object categories.
For a fair comparison, we use the same SSL experimental settings as STAC. When performing SSL experiments on the MS-COCO dataset, two experimental settings are used. In the first setting, only a small amount of data in the 118k labeled images is selected as the labeled set. The remainder is used as the unlabeled set. 
Under this setting, we are able to verify the performance of the SSL algorithm when there is only a small amount of labeled data. In the second setting, the entire 118k images are used as the labeled set and the additional 123k unlabeled images are used as the unlabeled set, which enables us to verify whether SSL algorithm can further improve the performance of the detector when large-scale labeled images already exist. In the first experimental setting, we randomly selected 1\%, 2\%, 5\%, and 10\% from the 118k labeled images as the labeled set.

In addition, we also test on PASCAL VOC~\cite{everingham2010pascal} following~\cite{jeong2019consistency} and report the mAP over 20 object categories. We use the trainval set of VOC07 as labeled data, which consists of $\sim$5k images, and the unlabeled data contains the trainval set of VOC12 ($\sim$11k images) and the subset of MS-COCO with the same classes as PASCAL VOC ($\sim$95k images). We evaluate the performance on the test set of VOC07 and report the mAPs at IoU=0.5, IoU=0.75, and IoU=0.5:0.95.

\begin{table*}[t!]
\begin{center}
    \resizebox{0.9\linewidth}{!}{
    \begin{tabular}{l||c|c|c|c|c|c}
    \toprule
    Methods     & Backbone & 1\% COCO      & 2\% COCO       & 5\% COCO       & 10\% COCO      & 100\% COCO \\ \midrule
    Supervised  & R50-FPN & 9.05$\pm$0.16 & 12.70$\pm$0.15 & 18.47$\pm$0.22 & 23.86$\pm$0.81 & 37.63      \\
    CSD$^{\dagger}$~\cite{jeong2019consistency} & R50-FPN & 10.20$\pm$0.15 (\textcolor{red}{+1.15}) & 13.60$\pm$0.10 (\textcolor{red}{+0.90}) & 18.90$\pm$0.10 (\textcolor{red}{+0.43}) & 24.50$\pm$0.15 (\textcolor{red}{+0.64}) & 38.87 (\textcolor{red}{+1.24}) \\
    STAC\cite{sohn2020simple} & R50-FPN & 13.97$\pm$0.35 (\textcolor{red}{+4.92}) & 18.25$\pm$0.25 (\textcolor{red}{+5.55}) & 24.38$\pm$0.12 (\textcolor{red}{+5.91}) & 28.64$\pm$0.21 (\textcolor{red}{+4.78}) & 39.21 (\textcolor{red}{+1.58}) \\ \midrule
    Instant-Teaching  (ours) & R50-FPN & \textbf{16.00$\pm$0.20} (\textcolor{red}{+6.95})  & \textbf{20.70$\pm$0.30} (\textcolor{red}{+8.00})   & \textbf{25.50$\pm$0.05} (\textcolor{red}{+7.03})  &  \textbf{29.45$\pm$0.15} (\textcolor{red}{+5.59})  &      \textbf{39.60} (\textcolor{red}{+1.97})      \\
    Instant-Teaching$^*$ (ours) & R50-FPN & \textbf{18.05$\pm$0.15} (\textcolor{red}{+9.00})  & \textbf{22.45$\pm$0.15} (\textcolor{red}{+9.75}) &  \textbf{26.75$\pm$0.05} (\textcolor{red}{+8.28})   & \textbf{30.40$\pm$0.05} (\textcolor{red}{+6.54}) &    \textbf{40.20} (\textcolor{red}{+2.57})        \\ \bottomrule
    \end{tabular}
    }
\end{center}
\vspace{-.1in}
\caption{Comparison of mAP for different semi-supervised methods on MS-COCO. CSD$^{\dagger}$ is our implementation of the CSD method based on the Faster-RCNN detector. Instant-Teaching$^*$ represents our Instant-Teaching framework with co-rectify scheme. The value in brackets represents the mAP improvement compared to the supervised model.}
\vspace{-.1in}
\label{tbl:result_coco}
\end{table*}

\subsection{Implementation details}


We implement our Instant-Teaching$^*$ framework based on the MMDetection toolbox~\cite{mmdetection}. To get a fair comparison, we follow STAC to use Faster-RCNN~\cite{shaoqing2017faster} with FPN~\cite{lin2017feature} as our object detector and use ResNet-50~\cite{he2016deep} as the feature extractor. The feature weights are initialized by the ImageNet-pretrained model.
Instant-Teaching$^*$ mainly contains three hyperparameters: $\lambda$, $\lambda_u$ and $\tau$, we set $\lambda=1.0$, $\lambda_u=1.0$ and $\tau=0.9$ unless otherwise specified.

All our experiments maintain the same training parameters as STAC. 
Specifically, we train the model using an SGD optimizer on 8 GPUs, with an initial learning rate of 0.01, a momentum of 0.9, a weight decay of 1$e-$4 and a total training step of 180k. The learning rate decays by 10$\times$ at 120k and 165k respectively. Moreover, we fix the mini-batch size to 16, in which the ratio between labeled images and unlabeled images is 1:1.
Following STAC, for 1\%, 2\%, 5\% and 10\% MS-COCO protocols, we use the \textbf{quick} learning schedule. For the 100\% protocol, we use the \textbf{standard} learning schedule.
The quick schedule adopts multi-scale training and the standard schedule adopts single-scale training, which is depicted in the Appendix A of STAC~\cite{sohn2020simple} and our supplementary materials.

\begin{table}[t!]
\begin{center}
    \resizebox{1\linewidth}{!}{
    \begin{tabular}{l||c|c|c|c|c}
    \toprule
    Methods   & Backbone  &  Unlabeled  &  AP$^{0.5:0.95}$     &  AP$^{0.5}$ & AP$^{0.75}$  \\ \midrule
    Supervised (Ours)                 & R50-FPN       &     &    43.60 & 76.70 & 44.50 \\ \midrule
    CSD~\cite{jeong2019consistency}   & R101-R-FCN        & \multirow{4}{*}{VOC12}  & - & 74.70 & - \\ 
    STAC~\cite{sohn2020simple}        & R50-FPN          &                         & 44.64 (\textcolor{red}{+1.04}) & 77.45 & - \\ 
    Instant-Teaching                  & R50-FPN        &                         & \textbf{48.70} (\textcolor{red}{+5.10})  &  \textbf{78.30} & \textbf{52.00} (\textcolor{red}{+7.50}) \\
    Instant-Teaching$^*$              & R50-FPN        &                         & \textbf{50.00} (\textcolor{red}{+6.40}) & \textbf{79.20}  & \textbf{54.00} (\textcolor{red}{+9.50}) \\ 
    \midrule
    CSD~\cite{jeong2019consistency}   & R101-R-FCN      & VOC12  & - & 75.10 & - \\
    STAC~\cite{sohn2020simple}        & R50-FPN        & \multirow{2}{*}{\&}                                 & 46.01 (\textcolor{red}{+2.41}) & 79.08 & - \\
    Instant-Teaching                  & R50-FPN        &                                 & \textbf{49.70} (\textcolor{red}{+6.10})  &  79.00 & \textbf{54.10} (\textcolor{red}{+9.60})\\
    Instant-Teaching$^*$              & R50-FPN        & COCO                                & \textbf{50.80} (\textcolor{red}{+7.20}) & \textbf{79.90} & \textbf{55.70} (\textcolor{red}{+11.20}) \\ \bottomrule
    \end{tabular}
    }
\end{center}
\vspace{-.1in}
\caption{Comparison of mAP for different semi-supervised methods on VOC07. We report the mAP at IoU=0.50:0.95 (AP$^{0.5:0.95}$), IoU=0.5 (AP$^{0.5}$) and IoU=0.75 (AP$^{0.75}$), which are the standard metrics for object detection~\cite{lin2014microsoft,cai2018cascade}.}
\label{tbl:result_voc}
\vspace{-.1in}
\end{table}

\subsection{Results}

We will make a detailed comparison with the supervised baseline and state-of-the-art SSOD methods, including CSD~\cite{jeong2019consistency} and STAC~\cite{sohn2020simple}. The detailed results are summarized in Table~\ref{tbl:result_coco} and Table~\ref{tbl:result_voc}.

As depicted in Table~\ref{tbl:result_coco}, our Instant-Teaching$^*$ outperforms state-of-the-art methods by a large margin under all experimental settings of the MS-COCO dataset. Specifically, for the 1\% protocol, Instant-Teaching$^*$ improves mAP from STAC's 13.97 to 18.05, which achieves 4.08 mAP improvement; for the 2\% protocol, Instant-Teaching$^*$ improves mAP from STAC's 18.25 to 22.45, which achieves 4.2 mAP improvement. Instant-Teaching$^*$ also brings significant improvement in mAP when there are more labeled data: 24.38 to 26.75 on the 5\% protocol, 28.64 to 30.40 on the 10\% protocol. For the 100\% protocol, our Instant-Teaching$^*$ still achieves about 1.0 mAP improvement under the high benchmark of 39.21 mAP.

We also observe a similar trend on PASCAL VOC experiments. 
As depicted in Table~\ref{tbl:result_voc},
when compared with STAC, with VOC07 as labeled data and VOC12 as unlabeled data, our Instant-Teaching$^*$ improves mAP from 44.64 to 50.00, which demonstrates 5.36 absolute mAP improvement. When there are more unlabeled data introduced (the subset of MS-COCO), Instant-Teaching$^*$ can further improve mAP from STAC's 46.01 to 50.80.
We also observe that the improvement of AP$^{0.75}$ of Instant-Teaching$^*$ is more prominent compared to that of AP$^{0.5}$. In other words, the improvement of mAP (AP$^{0.5:0.95}$) mainly comes from the improvement of predicted high-quality bounding boxes.
%
We also perform ablation studies on our Instant-Teaching$^*$ with different backbones in the Appendix (\ref{sec:backbone}), demonstrating the scalability of our method.

\section{Ablation Study}
 
\begin{figure}[t!]
	\begin{center}
	\includegraphics[scale=0.44]{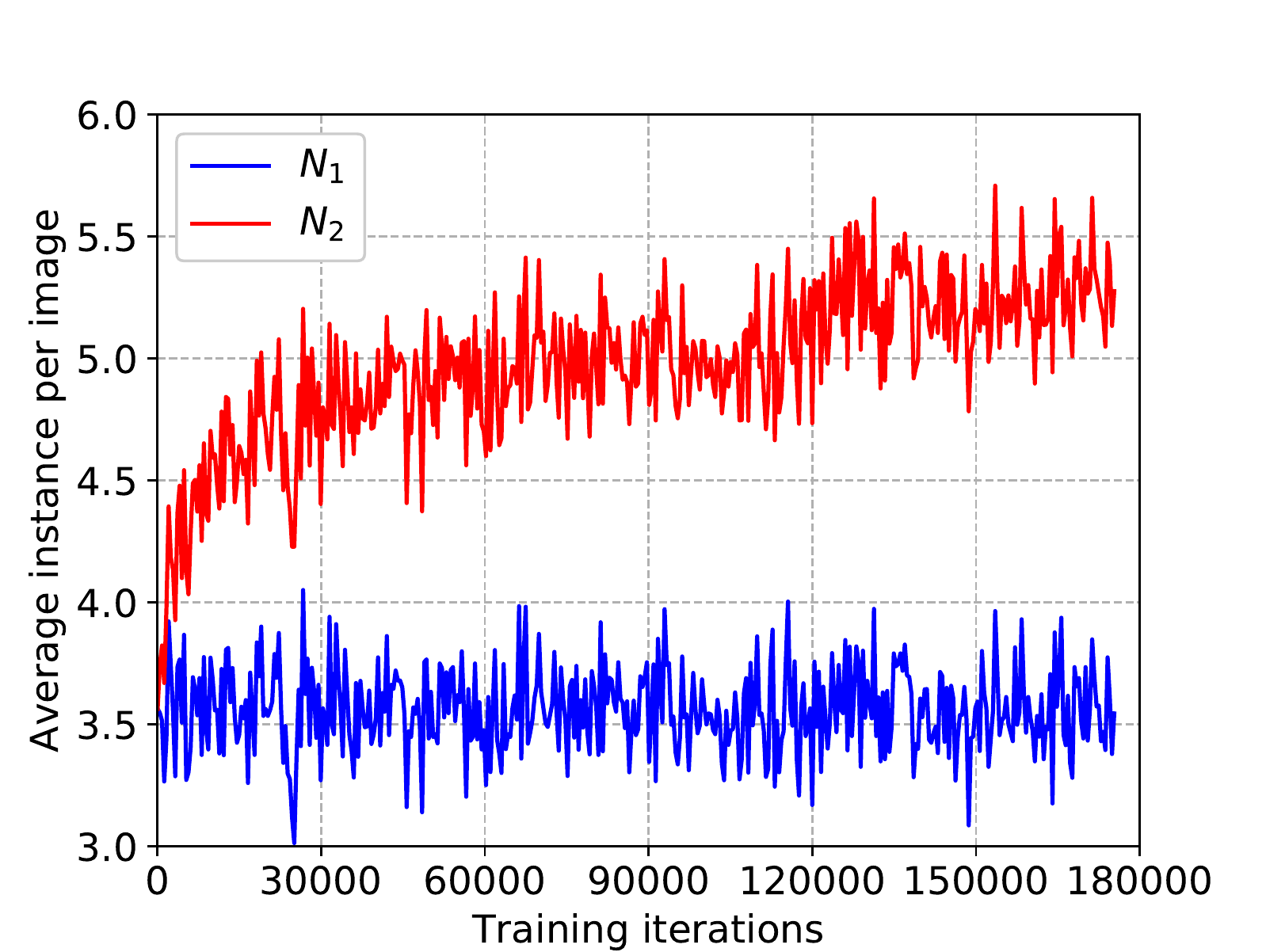}
	\end{center}
	\vspace{-.1in}
	\caption{Changes in the number of annotations per image during training. $N_1$ refers human-annotated instances and $N_2$ refers total instances including human-annotated and model generated.}
	\label{fig:online_pseudo}
	\vspace{-.25in}
\end{figure}

\subsection{Instant pseudo labeling}

As shown in Fig.~\ref{fig:online_pseudo}, we report the average number of annotated instances per image during each training iteration, in which $N_1$ denotes the number of only human-annotated instances and $N_2$ denotes the number of total instances including human-annotated and model generated (pseudo annotations).
It can be observed that the number of high-quality pseudo annotations ($N_2 - N_1$) gradually increases during the training process. Namely, as the model converges, the quantity of high-quality pseudo annotations can be instantly improved.


From Table~\ref{tbl:ablation_mm}, we can also observe that at the protocol of 5\% MS-COCO with 8$\times$ unlabeled data, Instant-Teaching improves mAP from STAC's 23.14 to 24.70 using only color jittering and Cutout~\cite{devries2017improved} as the strong data augmentations. Without using more strong data augmentations, our Instant-Teaching already outperforms the state-of-the-art STAC method.
These results prove that our instant pseudo labeling can finally achieve higher performance by continuously improving the pseudo annotations.


\begin{table}[t!]
\begin{center}
    \resizebox{1\linewidth}{!}{
	\begin{tabular}{l|c|c|c|c||c}
 	\toprule
	\multirow{2}{*}{Methods} & \multicolumn{4}{c||}{Strong data augmentations}    & \multirow{2}{*}{mAP}  \\ \cmidrule{2-5}
	                         & Color+Cutout & Geometric & Mixup & Mosaic              &  \\
	\midrule
	STAC\cite{sohn2020simple}           & $\surd$ & $\surd$  &    &            & 23.14  \\
	\midrule
	\multirow{5}{*}{Instant-Teaching}   & $\surd^{\star}$ &   &   &    & 21.60 (\textcolor{blue}{-1.54}) \\
	                                    & $\surd$ &   &           &            & 24.70 (\textcolor{red}{+1.56}) \\
	                                    & $\surd$ &   &  $\surd$  &            & 25.40 (\textcolor{red}{+2.26}) \\
                    	                & $\surd$ &   &           &  $\surd$   & 25.00 (\textcolor{red}{+1.86}) \\
	                                    & $\surd$ &   &  $\surd$  &  $\surd$   & \textbf{25.60} (\textcolor{red}{+2.46})  \\
 	\bottomrule
 	\end{tabular}
 	}
 \end{center}
 \vspace{-.1in}
 \caption{Comparison of mAP of Instant-Teaching trained with various data augmentation methods at the protocol of 5\% MS-COCO and 8$\times$ unlabeled data. $\surd^{\star}$ denotes that we also apply strong augmentations ``Color+Cutout'' to unlabeled data in the first step during instant pseudo labeling.}
 \label{tbl:ablation_mm}
\end{table}

\subsection{Strong data augmentation}

In weak-strong data augmentation scheme, the choice of strong augmentations directly affects the performance of the final SSOD model. We extend the strong augmentations of STAC from color jittering, geometric transformation and Cutout to include Mixup and Mosaic. Note that, we do not apply geometric transformation, mainly because the online geometric transformation of pseudo annotations is more complicated, and we leave it for future work.



As shown in Table~\ref{tbl:ablation_mm}, we first also apply strong augmentations (Color+Cutout) to unlabeled data in the first step during the pseudo labeling phase. This method gives us 1.54 mAP drop compared with STAC. The observation verifies our hypothesis, \ie, the key of weak-strong data augmentation scheme lies in the difference between weak augmentations and strong augmentations.
Furthermore, we find that using either Mixup or Mosaic can improve the performance of Instant-Teaching. Instant-Teaching can obtain the best performance by using Mixup and Mosaic data augmentations together, increasing mAP from 23.14 of STAC to 25.60. These observations indicate that our extended weak-strong data augmentations can further improve the performance of SSOD.

Note that we only use Mixup and Mosaic data augmentations for unlabeled data for a fair comparison with STAC.

\begin{table}[t!]
\begin{center}
    \resizebox{1\linewidth}{!}{
	\begin{tabular}{l||c|c|c|c|c|c}
 	\toprule
	\multirow{2}{*}{Methods} & \multirow{2}{*}{Labeled Size} & \multicolumn{5}{c}{Unlabeled Size} \\ \cline{3-7}
	&  & 1$\times$ & 2$\times$ & 4$\times$ & 8$\times$ & Full \\
	\midrule
	STAC\cite{sohn2020simple} & \multirow{2}{*}{5\% COCO} & 19.81 & 20.79 & 22.09 & 23.14 & 24.38$\pm$0.12 \\
	Instant-Teaching &  &  \textbf{23.60}  & \textbf{24.30}  & \textbf{25.30}  & \textbf{25.60}  & \textbf{25.60$\pm$0.14} \\
	\midrule
	STAC\cite{sohn2020simple} & \multirow{2}{*}{10\% COCO} & 25.38 & 26.52 & 27.33 & 27.95 & 28.64$\pm$0.21 \\
	Instant-Teaching & & \textbf{28.80} & \textbf{29.00} & \textbf{29.20} & \textbf{29.50} & \textbf{29.53$\pm$0.17} \\
 	\bottomrule
 	\end{tabular}
 	}
 \end{center}
 \vspace{-.1in}
 \caption{Comparison of mAP of Instant-Teaching trained with various scales of unlabeled data on MS-COCO. $\lbrack$$n$$\rbrack$$\times$ denotes the scale of unlabeled data is $\lbrack$$n$$\rbrack$ times larger than that of labeled data.}
 \label{tbl:ablation_2}
  \vspace{-.1in}
 \end{table}

\subsection{Size of unlabeled data}

In the field of semi-supervised object detection, the importance of the size of unlabeled data should not be ignored. Therefore in this section, we evaluate our method with 5\% and 10\% labeled data of MS-COCO while varying the size of unlabeled data from 1, 2, 4, and 8 times to that of the labeled data. The results are given in Table~\ref{tbl:ablation_2}.
We can observe that our method outperforms the state-of-the-art method STAC on all scales of unlabeled data. It is worth mentioning that, for both 5\% and 10\% labeled data, our Instant-Teaching method trained on 1$\times$ unlabeled data achieves 23.60 and 28.80 mAP respectively, which are even higher than STAC trained on 8$\times$ unlabeled data (23.14 and 27.95). This demonstrates that Instant-Teaching can efficiently leverage the unlabeled data.

\begin{figure}[t!]
	\begin{center}
	\includegraphics[scale=0.44]{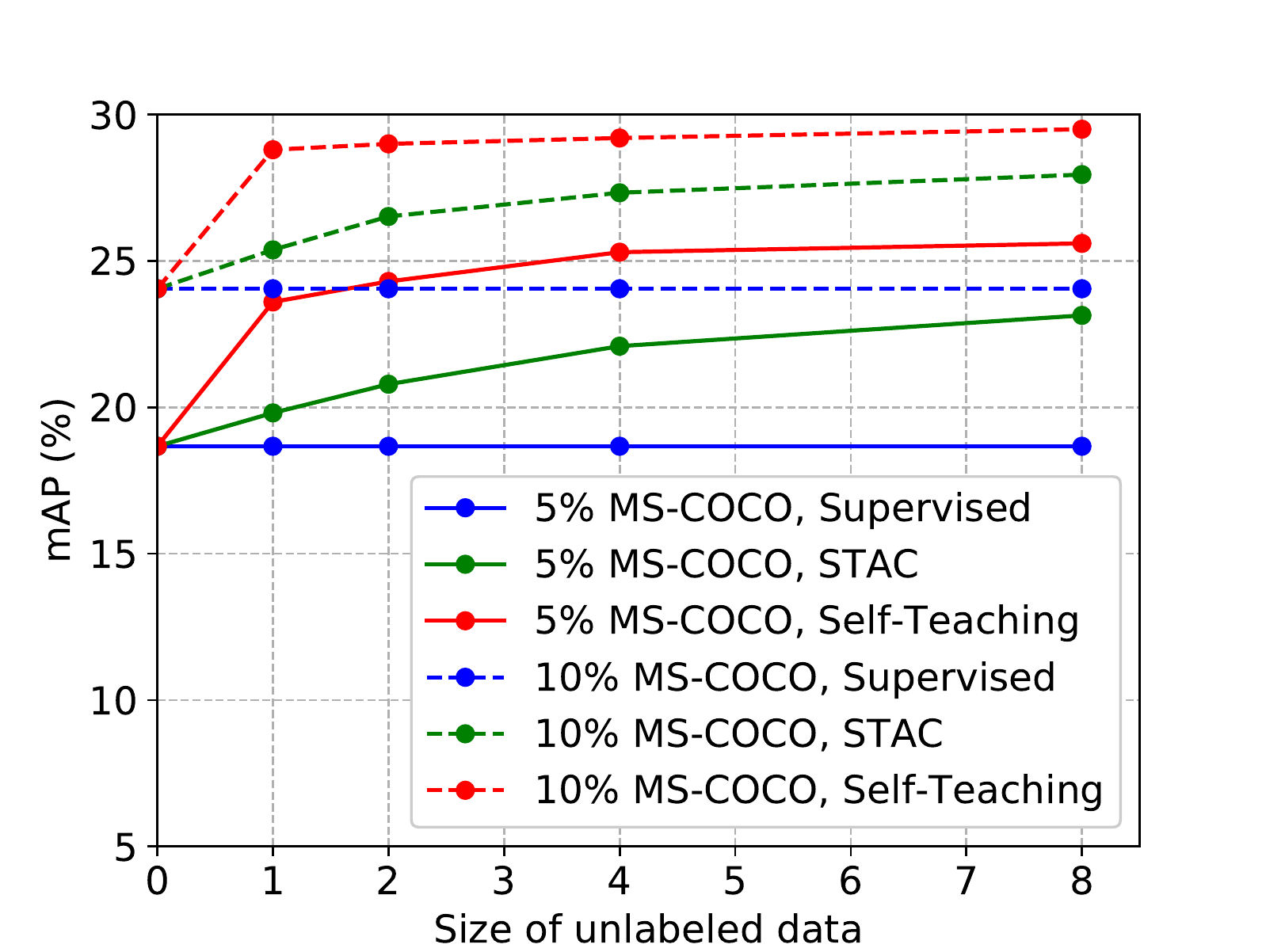}
	\end{center}
	\vspace{-.1in}
	\caption{Comparison of mAP w.r.t. the size of unlabeled data.}
	\label{fig:unlabel_size}
\end{figure}

From Fig.~\ref{fig:unlabel_size} we can observe that our Instant-Teaching (without co-rectify) outperforms the supervised model and the state-of-the-art method STAC by a large margin. 
We also find that as the size of unlabeled data increases, both STAC and Instant-Teaching suffer a ``ceiling effect'': as the performance gets closer to the ceiling, the improvement becomes smaller.

\begin{table}[t!]
\begin{center}
    \resizebox{0.65\linewidth}{!}{
    \begin{tabular}{l||c|c|c|c}
    \toprule
    $\tau$     & 0.3      & 0.5       & 0.7       & 0.9      \\ \midrule
    mAP (\%)   & 26.30    & 27.70     & 28.70     & 29.80  \\ \bottomrule
    \end{tabular}
    }
\end{center}
\vspace{-.1in}
\caption{Comparison of mAP with various values of confidence threshold $\tau$.}
\vspace{-.1in}
\label{tbl:tau}
\end{table}

\subsection{Analysis of $\tau$ and $\lambda_u$}

We analyze the effect of the confidence threshold $\tau$ and the unsupervised loss weight $\lambda_u$ in this section. Our Instant-Teaching method is tested with 10\% MS-COCO as labeled data and the remainder as unlabeled data. We first analyze the effect of $\tau$. As shown in Table~\ref{tbl:tau}, we test Instant-Teaching with $\lambda_u=1.0$ and $\tau \in \lbrace0.3, 0.5, 0.7, 0.9\rbrace$. The result shows that the model can achieve better performance by varying the threshold value $\tau$ from 0.3 to 0.9, which indicates $\tau=0.9$ is a better choice to select high-quality pseudo annotations for unlabeled data.

\begin{figure}[t!]
	\begin{center}
	\includegraphics[scale=0.44]{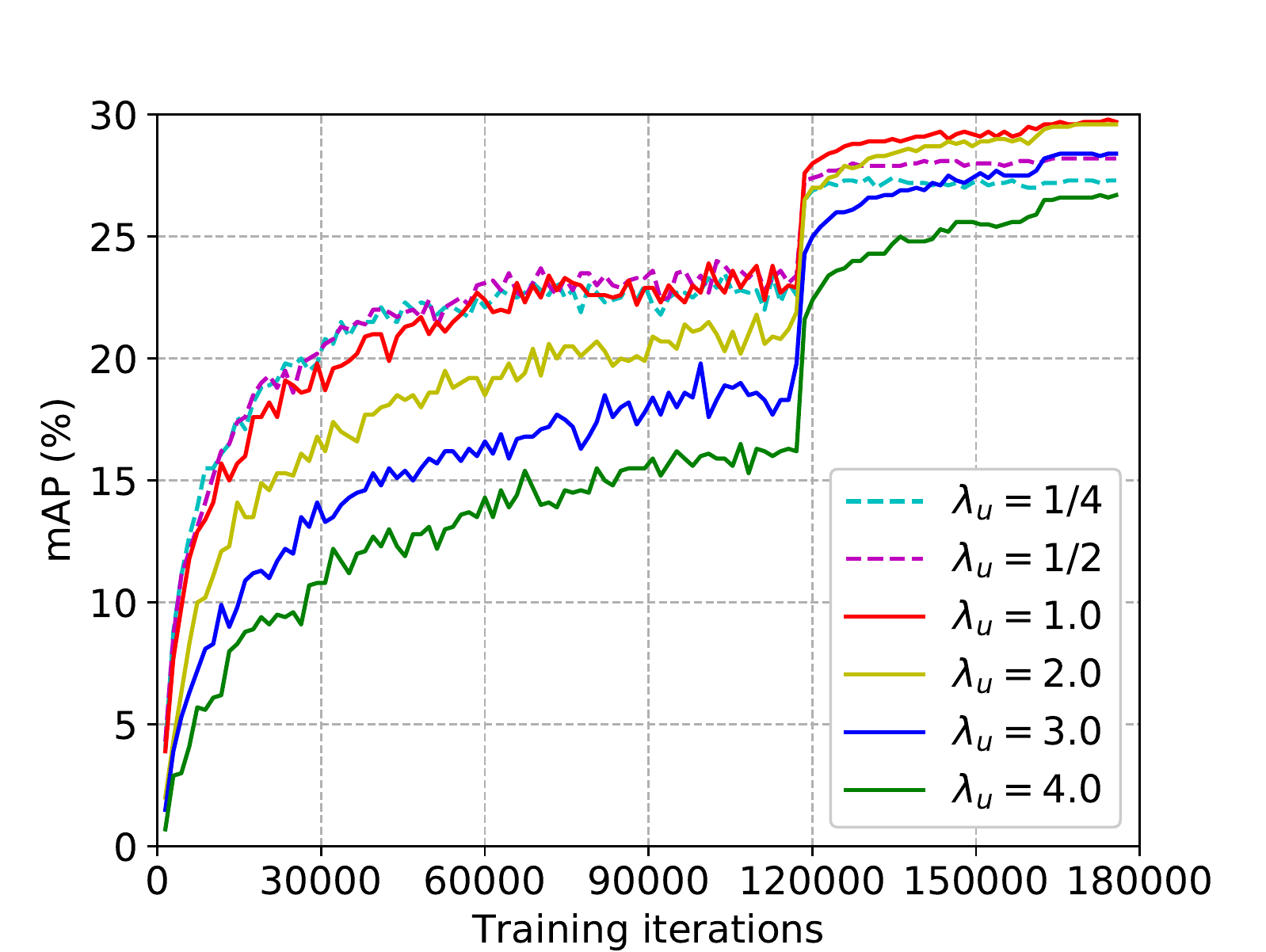}
	\end{center}
	\vspace{-.1in}
	\caption{Comparison of mAP with various values of $\lambda_u$ along training iterations.}
	\label{fig:lambda_u}
	\vspace{-.15in}
\end{figure}

When analyzing the effect of unsupervised loss weight $\lambda_u$, we fix $\tau=0.9$ and vary the value of $\lambda_u$ from 1/4 to 4. 
As can be seen in Fig.~\ref{fig:lambda_u}, Instant-Teaching achieves the best performance when $\lambda_u = 1.0$ and the mAP only slightly drop when $\lambda_u$ becomes larger or smaller, which indicates that Instant-Teaching is 
relatively robust to $\lambda_u$. 

We can also observe that Instant-Teaching achieves a higher mAP with a smaller value of $\lambda_u$ (\eg, 1/4, 1/2) during the early training iterations.
In other words, in the early stages of training, the quality (quantity) of pseudo annotations is low, and the model should pay more attention to the labeled data. In this paper, we use a constant $\lambda_u$, and take the dynamic adjustment of $\lambda_u$ as future work.

\begin{figure}[t!]
	\begin{center}
	\includegraphics[scale=0.44]{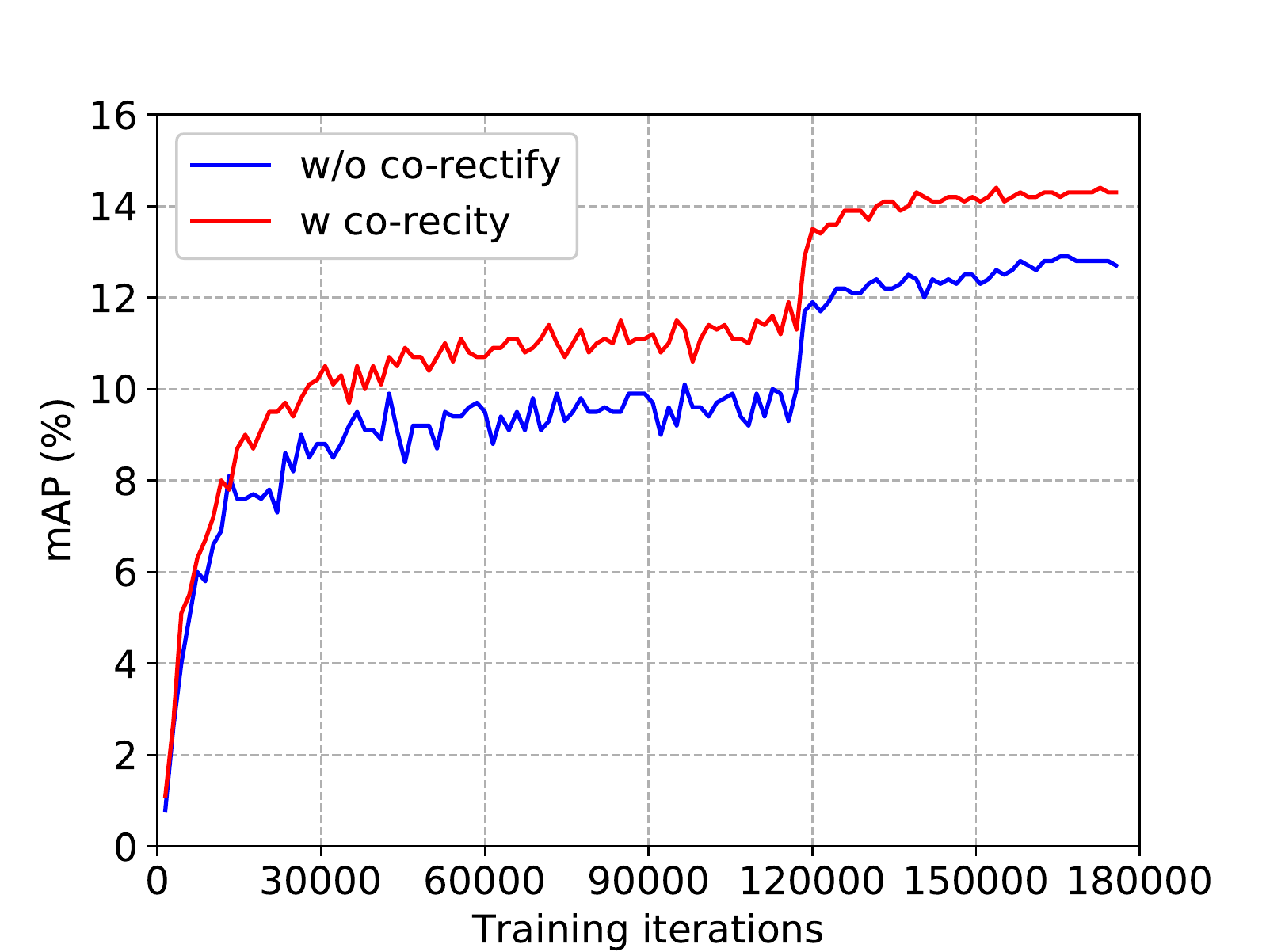}
	\end{center}
	\vspace{-.1in}
	\caption{Comparison of mAP of generated pseudo annotations with different training iterations. The model is trained based on Instant-Teaching with and without co-rectify respectively.}
	\label{fig:rectify_map}
	\vspace{-.15in}
\end{figure}

\subsection{Analysis of co-rectify}

We further propose a co-rectify scheme based on Instant-Teaching to alleviate the confirmation bias problem in SSL, which is shown in Fig.~\ref{fig:ssl_framework} (Instant-Teaching$^*$).
We analyze the effect of co-rectify using 1\% labeled data and the remaining 99\% as unlabeled data (1\% MS-COCO protocol). The model is trained based on Instant-Teaching with and without our co-rectify scheme respectively. For evaluation, we test on 5k labeled data, which is randomly selected from the 99\% unlabeled data of MS-COCO.

Note that, to verify whether the co-rectify scheme is able to generate more high-quality pseudo annotations, we compare the mAP of predicted pseudo annotations with score larger than 0.9 (same as $\tau$ during training).
As shown in Fig.~\ref{fig:rectify_map}, we can directly observe that the model trained with co-rectify scheme obtains better performance faster, and is able to consistently improve the performance of our Instant-Teaching along the training iterations. 

\begin{figure}[t!]
	\begin{center}
	\includegraphics[width=0.47\textwidth]{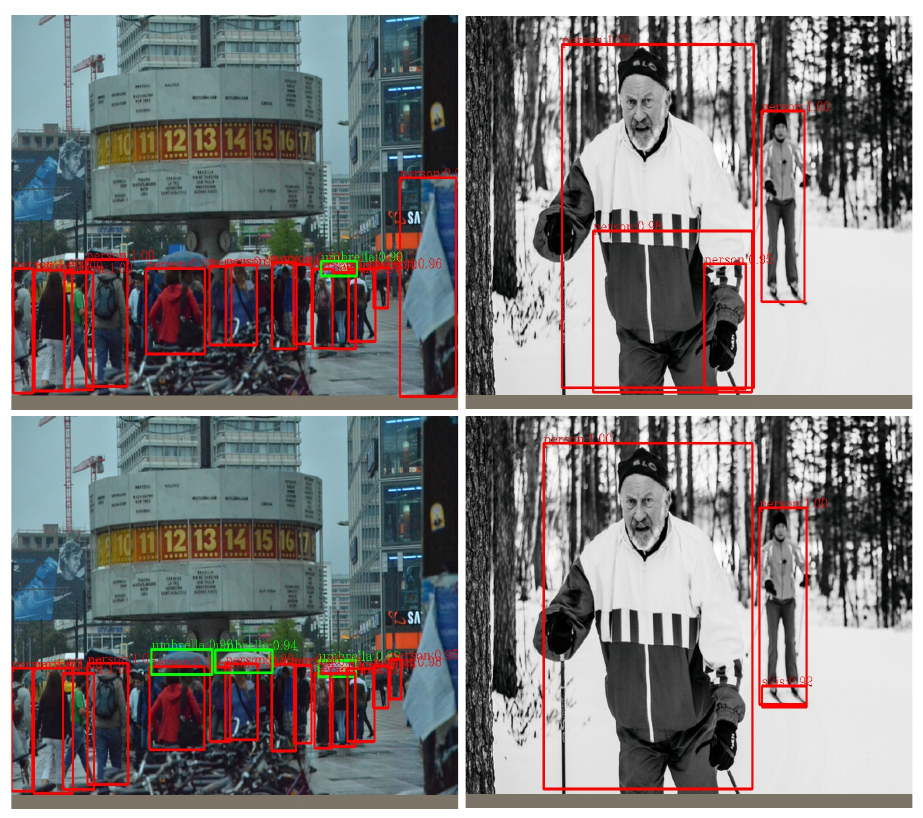}
	\end{center}
	\vspace{-.1in}
	\caption{Visualization of predicted pseudo annotations whose confidence scores are larger than 0.9 for unlabeled data. The first row denotes the results of Instant-Teaching (without co-rectify) and the second row denotes the results of Instant-Teaching$^*$.}
	\label{fig:rectify_vis}
	\vspace{-.15in}
\end{figure}

In addition, we visualize the pseudo annotations for some unlabeled data in Fig.~\ref{fig:rectify_vis}. The results are generated at the same training iteration (120k) with and without the co-rectify scheme respectively. We can observe that Instant-Teaching cooperated with co-rectify scheme can filter out some false predictions and generate more high-quality pseudo annotations at the same time. In summary, the co-rectify scheme is able to alleviate the confirmation bias problem and further improve the performance of Instant-Teaching.

\section{Conclusion}

In this paper, we revisit semi-supervised object detection (SSOD) and propose a simple and effective end-to-end SSOD framework --- Instant-Teaching, which uses instant pseudo labeling with extended weak-strong data augmentations for teaching during each training iteration.
Based on Instant-Teaching, we further propose a co-rectify scheme to alleviate the confirmation bias problem and further improve the performance.
Extensive experiments on MS-COCO and PASCAL VOC demonstrate the significant superiority of our method.
Although we evaluate with the two-stage detector Faster-RCNN~\cite{shaoqing2017faster}, our proposed Instant-Teaching$^*$ is a general SSOD framework and is not restricted to the object detection models. This means Instant-Teaching$^*$ can be directly applied to other detectors, \eg, one-stage detectors SSD~\cite{liu2016ssd} and FCOS~\cite{tian2019fcos}, which we will leave for future work.

{\small
\bibliographystyle{ieee_fullname}
\bibliography{egbib}
}

\appendix
\section{Appendix}
\subsection{Learning Schedules}
We provide more details on different learning schedules used in our experiments.

\subsubsection{MS-COCO: Quick}
\paragraph{Training}
\begin{itemize}
\item[-] \textbf{Batch size:} 16. 
\item[-] \textbf{LR decay:} [0.01 ($\le$120k), 0.001 ($\le$160k), 0.0001 ($\le$180k)].
\item[-] \textbf{Data processing:} Short edge size is sampled between 500 and 800 if the long edge is less than 1024 after resizing.
\item[-] \textbf{Batch per image for training Faster-RCNN head:} 64.
\end{itemize}

\paragraph{Testing}
\begin{itemize}
\item[-] \textbf{Data processing:} Short edge size is fixed to 800 if the long edge is less than 1024 after resizing.
\item[-] \textbf{Score threshold for testing Faster-RCNN head:} 0.001.
\end{itemize}

\subsubsection{MS-COCO: Standard, [$n$] $\times$}
\paragraph{Training}
\begin{itemize}
\item[-] \textbf{Batch size:} 16. 

\item[-] \textbf{LR decay (1$\times$):} [0.01 ($\le$120k), 0.001 ($\le$160k), 0.0001 ($\le$180k)].

\item[-] \textbf{LR decay (2$\times$):} [0.01 ($\le$240k), 0.001 ($\le$320k), 0.0001 ($\le$360k)].

\item[-] \textbf{LR decay (3$\times$):} [0.01 ($\le$420k), 0.001 ($\le$500k), 0.0001 ($\le$540k)].

\item[-] \textbf{Data processing:} Short edge size is fixed to 800 if the long edge is less than 1333 after resizing.

\item[-] \textbf{Batch per image for training Faster-RCNN head:} 512.
\end{itemize}

\paragraph{Testing}
\begin{itemize}
\item[-] \textbf{Data processing:} Short edge size is fixed to 800 if the long edge is less than 1333 after resizing.
\item[-] \textbf{Score threshold for testing Faster-RCNN head:} 0.001.
\end{itemize}

\subsubsection{PASCAL VOC}
\paragraph{Training}
\begin{itemize}
\item[-] \textbf{Batch size:} 16. 

\item[-] \textbf{LR decay:} [0.01 ($\le$120k), 0.001 ($\le$160k), 0.0001 ($\le$180k)].
\item[-] \textbf{Data processing:} Short edge size is fixed to 600 if the long edge is less than 1000 after resizing. 
\item[-] \textbf{Batch per image for training Faster-RCNN head:} 256.
\end{itemize}

\paragraph{Testing}
\begin{itemize}
\item[-] \textbf{Data processing:} Short edge size is fixed to 600 if the long edge is less than 1000 after resizing. 
\item[-] \textbf{Score threshold for testing Faster-RCNN head:} 0.001.
\end{itemize}

\subsection{Hyperparameters}
In this section, we provide descriptions of the hyperparameters used in our experiments, as shown in Tabel~\ref{tbl:hyper}. Unless otherwise specified, we use the same hyperparameters in both the MS-COCO and PASCAL VOC experiments.

\begin{table}[h]
\begin{center}
    \resizebox{0.48\textwidth}{!}{
    \begin{tabular}{l||c|c}
    \toprule
    Hyperparameters     & Description      &  Value   \\ \midrule
    $\lambda$           & The bounding box regression loss weight     & 1.0     \\
    $\lambda_{u}$       & The unsupervised loss weight    &  1.0  \\ 
    $\tau$              & The confidence threshold    &  0.9  \\ 
    $\alpha_{m}$        & The coefficient of $Beta$ distribution    &  1.0  \\ 
    $\lambda_{m}$       & The mixing coefficient of Mixup & $Beta(\alpha_{m}, \alpha_{m})$ \\
    LR         & The initial learning rate    &  0.01  \\ 
    Momentum   & The momentum used in SGD    &  0.9  \\ 
    Weight decay   & The weight decay    &  1$e-$4  \\ 
    Training steps   & The total training steps    &  180k  \\ 
    Batch size   & The batch size    &  16  \\ 
    Batch ratio        & The ratio between labeled and unlabeled images in a batch             &  1:1  \\ 
    \bottomrule
    \end{tabular}
    }
\end{center}
\caption{Descriptions of the hyperparameters used in our experiments.}
\label{tbl:hyper}
\end{table}

\begin{figure}[t!]
	\begin{center}
	\includegraphics[width=1\linewidth]{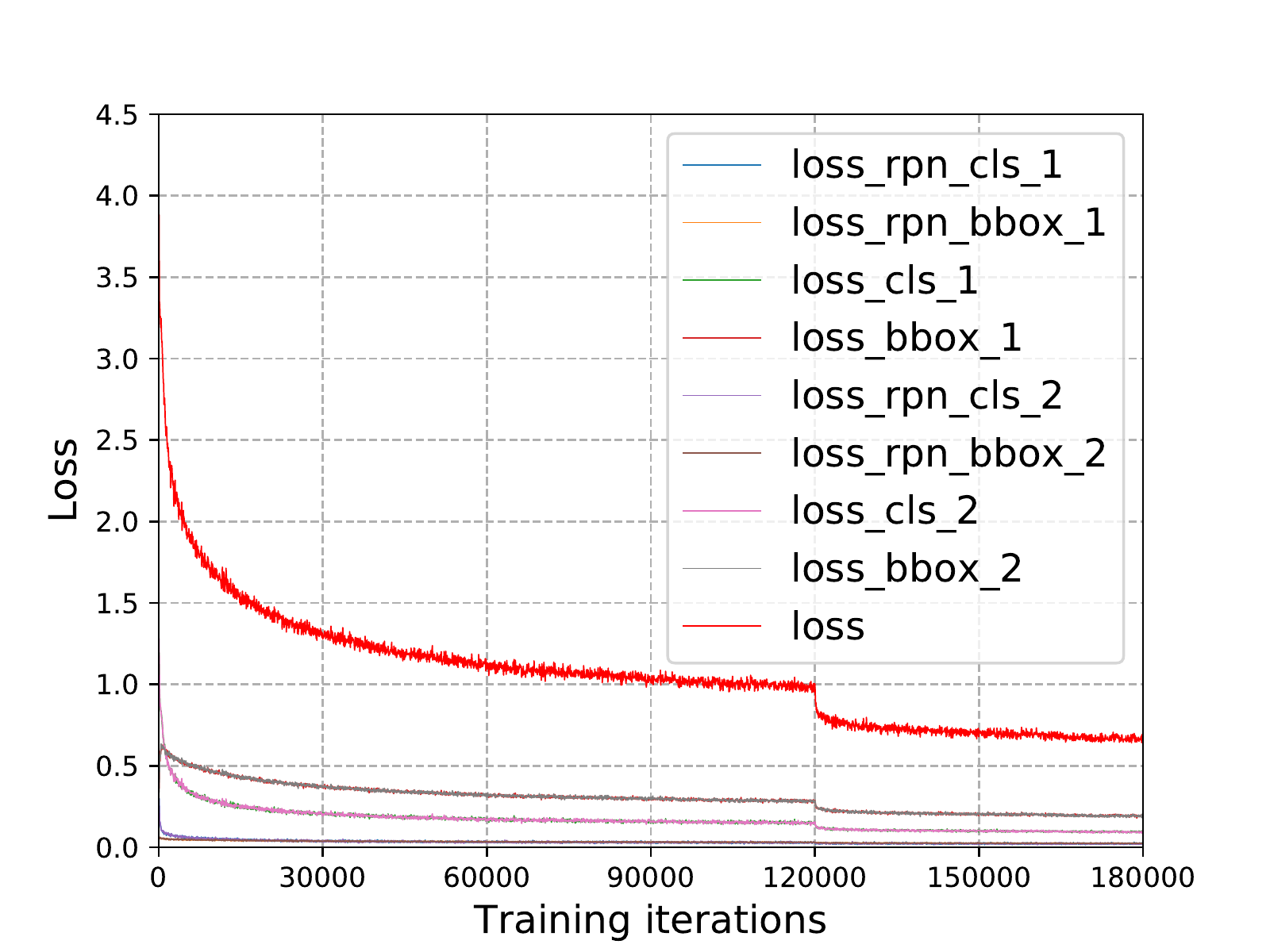}
	\end{center}
	\vspace{-.1in}
	\caption{Value change of loss w.r.t. training iterations.}
	\label{fig:convergence}
\end{figure}

\subsection{Convergence Analysis}
In this section, we empirically evaluate the convergence of Instant-Teaching$^*$. As shown in Fig.~\ref{fig:convergence}, we can observe that both the two models trained with our Instant-Teaching$^*$ method can reach a steady convergence. These results demonstrate that our Instant-Teaching$^*$ can not only achieve state-of-the-art results, it can also be trained easily.

\begin{table}[t!]
\centering
    \resizebox{0.9\linewidth}{!}{
    \begin{tabular}{l||c|c}
    \toprule
    Methods                 & Backbone                  &  2\% COCO    \\ \midrule
    Supervised              & \multirow{3}{*}{R50-FPN}  &  12.70$\pm$0.15   \\
    Instant-Teaching        &                           &  20.70$\pm$0.30 (\textcolor{red}{+8.00})      \\
    Instant-Teaching$^*$    &                           &  22.45$\pm$0.15 (\textcolor{red}{+9.75})      \\ 
    \midrule
    Supervised              & \multirow{3}{*}{R101-FPN}     &  15.80$\pm$0.50   \\
    Instant-Teaching        &                               &  22.10$\pm$0.15 (\textcolor{red}{+6.30})      \\
    Instant-Teaching$^*$    &                               &  23.50$\pm$0.20 (\textcolor{red}{+7.70})      \\ 
    \midrule
    Supervised              & \multirow{3}{*}{X101-32x4d-FPN}     &  16.60$\pm$0.20   \\
    Instant-Teaching        &                                     &  22.40$\pm$0.15 (\textcolor{red}{+5.80})      \\
    Instant-Teaching$^*$    &                                     &  24.20$\pm$0.15 (\textcolor{red}{+7.60})      \\
    \midrule
    Supervised              & \multirow{3}{*}{R2N-101-FPN}        &  17.4$\pm$0.30    \\
    Instant-Teaching        &                                     &  23.5$\pm$0.15 (\textcolor{red}{+6.10})     \\
    Instant-Teaching$^*$    &                                     &  25.9$\pm$0.20 (\textcolor{red}{+8.50})      \\
    \bottomrule
    \end{tabular}
    }
\caption{Comparison of mAP for different semi-supervised methods with different backbones on the 2\% MS-COCO protocol. The value in brackets represents the mAP improvement compared to the corresponding supervised model.}
\label{tbl:ablation_backbone}
\end{table}

\subsection{Ablation Study on Backbone}
\label{sec:backbone}
In this section, we verify the effect of different backbones on our Instant-Teaching$^*$ framework. From Table~\ref{tbl:ablation_backbone}, we replace the ResNet-50 backbone with ResNet-101 and test the efficacy of the supervised baseline, Instant-Teaching, and Instant-Teaching$^*$ method on the 2\% protocol respectively. We can observe that our Instant-Teaching$^*$ can reach better performance with a more powerful backbone. In other words, it is easy to elevate the performance of our Instant-Teaching$^*$ framework by using a more powerful backbone.

\end{document}